\documentclass[letterpaper]{article} 
\usepackage{aaai25}  
\usepackage{times}  
\usepackage{helvet}  
\usepackage{courier}  
\usepackage[hyphens]{url}  
\usepackage{graphicx} 
\usepackage{natbib}  
\usepackage{caption} 
\usepackage{algorithm}
\usepackage{algorithmic}
\usepackage{amsmath}
\usepackage{listings}
\usepackage{amssymb}
\usepackage{newfloat}
\usepackage{listings}

\title{G-VEval: A Versatile Metric for Evaluating \\ Image and
Video Captions Using GPT-4o}
\author{
    Tony Cheng Tong\textsuperscript{\rm 1}\equalcontrib,
    Sirui He\textsuperscript{\rm 2}\equalcontrib,
    Zhiwen Shao\textsuperscript{\rm 1,\rm 3}\thanks{Corresponding authors.},
    Dit-Yan Yeung\textsuperscript{\rm 1}\footnotemark[2]
}
\affiliations{
    \textsuperscript{\rm 1}The Hong Kong University of Science and Technology\\
    \textsuperscript{\rm 2}National University of Singapore\\
    \textsuperscript{\rm 3}China University of Mining and Technology\\
    ztangaj@connect.ust.hk, he.sirui@u.nus.edu, zhiwen@ust.hk, dyyeung@cse.ust.hk
}

\begin{document}

\maketitle

\begin{abstract}
Evaluation metric of visual captioning is important yet not thoroughly explored. Traditional metrics like BLEU, METEOR, CIDEr, and ROUGE often miss semantic depth, while trained metrics such as CLIP-Score, PAC-S, and Polos are limited in zero-shot scenarios. Advanced Language Model-based metrics also struggle with aligning to nuanced human preferences. To address these issues, we introduce G-VEval, a novel metric inspired by G-Eval and powered by the new GPT-4o. G-VEval uses chain-of-thought reasoning in large multimodal models and supports three modes: reference-free, reference-only, and combined, accommodating both video and image inputs. We also propose MSVD-Eval, a new dataset for video captioning evaluation, to establish a more transparent and consistent framework for both human experts and evaluation metrics. It is designed to address the lack of clear criteria in existing datasets by introducing distinct dimensions of Accuracy, Completeness, Conciseness, and Relevance (ACCR). Extensive results show that G-VEval outperforms existing methods in correlation with human annotations, as measured by Kendall tau-b and Kendall tau-c. This provides a flexible solution for diverse captioning tasks and suggests a straightforward yet effective approach for large language models to understand video content, paving the way for advancements in automated captioning.
\end{abstract}

\begin{links}
    \link{Code}{https://github.com/ztangaj/gveval}
    \link{Published version}{TBD}
\end{links}

\lstset{basicstyle=\ttfamily, numbers=none}

\section{ Introduction}
\label{sec:intro}

Visual captioning, the task of generating descriptive text from visual content, represents a crucial intersection between computer vision and natural language processing. This field primarily addresses the complex challenge of enabling machines to interpret and articulate visual data. Recently, researchers have integrated large language models (LLMs) to enhance the capabilities of visual captioning systems, resulting in more precise and generalizable captions that benefit numerous downstream applications. These advanced systems, known as large vision-language models (LVLMs), now assist the visually impaired, improve educational technologies, and enhance the autonomy of robotics \cite{li2020oscar, zhang2021vinvl}. 

Despite these advancements, evaluating V-LLMs remains challenging. Traditional metrics like BLEU, METEOR, CIDEr, and ROUGE often miss the semantic depth of captions. Trained metrics such as CLIP-Score and PAC-S, and Polos offer better language understanding but are limited in zero-shot scenarios. Additionally, metrics leveraging LLMs, like CLAIR, show strong human correlation but face limitations in interpretability and applicability to tasks beyond reference-only evaluation.

\begin{figure}
\centering
\includegraphics[width=\linewidth]{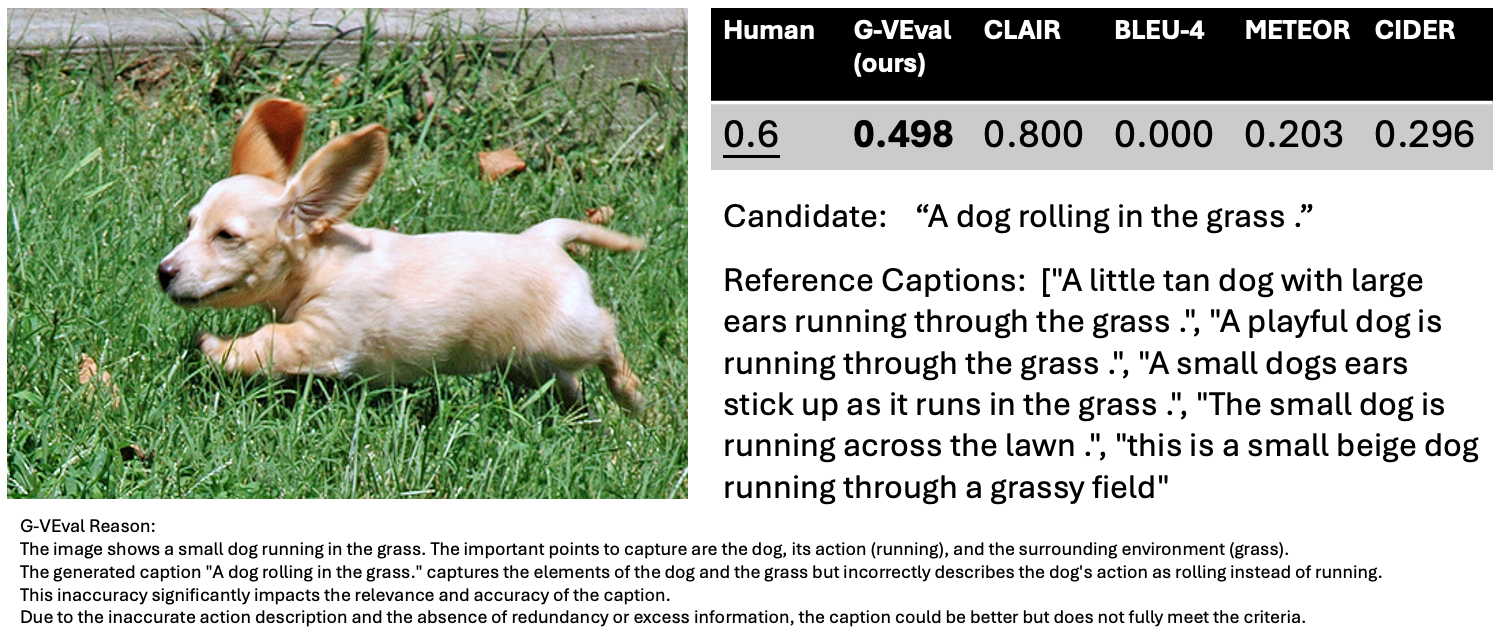}
\caption{Example of caption evaluation.}
\label{fig:sample-dog}
\end{figure}

To address these challenges, we introduce G-VEval, a novel metric inspired by G-Eval~\cite{liu2023geval} in natural language generation (NLG). G-Eval pioneered the use of GPT-4 for evaluation by leveraging chain-of-thought reasoning and addressing the probabilistic nature of model outputs through the calculation of expected values. Building on these innovations, G-VEval incorporates genuine chain-of-thought reasoning within large multimodal models and extends this approach to visual content, supporting three evaluation modes: reference-free, reference-only, and combined, making it applicable to both image and short video captioning. Additionally, we propose MSVD-Eval, a new dataset designed to address the limitations of existing evaluation datasets by providing clear criteria for assessing video captions across four dimensions: Accuracy, Completeness, Conciseness, and Relevance (ACCR).

G-VEval aims to bridge the gaps left by existing metrics like CLAIR and G-Eval by integrating the strengths of both, while also extending the evaluation framework to video captioning. By leveraging in-context chain-of-thought reasoning and multimodal capabilities, G-VEval delivers consistent and high-quality evaluations across a range of captioning tasks. An example of G-VEval evaluation result is shown in Figure \ref{fig:sample-dog}. Our results demonstrate that G-VEval achieves superior correlation with human judgments compared to existing methods, offering a robust and adaptable evaluation framework for future research in automated captioning.

\section{Related Work}
\label{sec:related_work}

In this section, we review existing metrics for evaluating image and video captioning, categorizing them into untrained metrics, trained metrics, and advanced language model-based metrics. Our goal is to highlight the strengths and limitations of each approach, ultimately establishing the need for a more versatile and robust metric like G-VEval.

\subsection{ Untrained Metrics}
Untrained metrics primarily rely on n-gram matching between generated captions and reference captions. These metrics, including BLEU, ROUGE, METEOR, and CIDEr, are popular due to their simplicity and ease of implementation.

BLEU calculates the precision of n-grams in the generated caption against reference captions \cite{papineni2002bleu}. While widely used in both machine translation and image captioning, BLEU struggles with synonyms and varied sentence structures, which can lead to lower scores for high-quality captions generated by advanced models like V-LLMs. ROUGE focuses on recall by comparing overlapping n-grams, word sequences, and word pairs between the generated and reference captions \cite{lin2004rouge}. This metric is commonly used in text summarization and image captioning but shares similar limitations with BLEU regarding semantic depth and synonym handling. METEOR combines precision and recall while incorporating synonym matching, stemming, and paraphrase detection \cite{banerjee2005meteor}. It offers a more nuanced evaluation compared to BLEU and ROUGE but still relies heavily on word-level matches.

CIDEr uses TF-IDF weighting for n-grams, emphasizing consensus among multiple references \cite{vedantam2015cider}. It is specifically designed for image captioning evaluation but can falter when the generated captions use different wording than the references, especially with synonyms. Despite their widespread use, these untrained metrics often fail to align well with human judgment, particularly for captions generated by models that use advanced language understanding, such as V-LLMs. While these metrics can be applied to video captioning tasks, they do not account for visual content, limiting their effectiveness in this domain.

\subsection{ Trained Metrics}
Trained metrics leverage pre-trained embeddings or human-labeled data, offering greater flexibility in language understanding. Embedding-based metrics, such as BERTScore, evaluate the similarity between generated and reference captions using contextual embeddings from BERT \cite{zhang2019bertscore}. MoverScore enhances BERTScore with soft alignments and advanced aggregation methods \cite{zhao2019moverscore}. These metrics are more flexible with synonyms and sentence segmentation but do not consider visual content, which limits their alignment with human preferences.

To address this gap, researchers have developed metrics based on the cross-modal embeddings of vision-language models. CLIP-Score measures the similarity between generated captions and image content using the CLIP model \cite{hessel2021clipscore}. CLIP-ViT-B-32 and CLIP-ViT-L-14 are commonly used versions, encoding images into 512-dimensional and 768-dimensional vectors, respectively. However, CLIP's encoding lacks the detail needed for fine-grained visual captioning, and its applicability to video captioning is limited. The only notable work in this area is EMScore, which evaluates video captioning via coarse-grained and fine-grained embedding matching \cite{shi2022emscore}.

Supervised metrics, such as PAC-S and Polos, are trained on datasets derived from human evaluations, showing high correlation with human preferences. PAC-S uses contrastive learning and human-labeled data to evaluate captions, emphasizing positive augmentation \cite{sarto2023positiveaugmented}. Polos, developed using multimodal metric learning from human feedback, is effective in aligning with human judgments \cite{wada2024polos}. However, their dependence on training data can lead to weak performance in zero-shot settings, limiting their broader applicability across diverse datasets.

\subsection{Advanced Language Model-Based Metrics}
Advanced language model-based metrics leverage the capabilities of large language models to provide more robust evaluations. CLAIR is an example of such a metric, using LLMs with simple prompts to evaluate image captions \cite{chan2023clair}. While CLAIR shows strong performance in human correlation, it is limited to reference-only evaluation and lacks interpretability due to its reliance on simple prompts. Additionally, CLAIR has not been extended to video captioning tasks, which restricts its broader applicability.

G-Eval, although not a visual captioning evaluation metric, presents a more structured approach to utilizing LLMs for evaluation tasks, specifically in the context of summarization \cite{liu2023geval}. Unlike CLAIR, G-Eval calculates the expected value of the output from LLMs such as GPT-3 and GPT-4, addressing the challenges associated with the probabilistic nature of LLMs. While G-Eval claims to use chain-of-thoughts (CoT) reasoning by including evaluation steps in the prompt, it often produces single-digit outputs, lacking a genuine in-context reasoning process, which may limit the effectiveness of CoT.

Most recently, FLEUR proposes a novel evaluation framework by utilizing LLaVA, a vision-language model that can directly assess image-caption pairs \cite{lee-etal-2024-fleur}. Unlike CLAIR's reference-dependent approach, FLEUR enables direct comparison between images and captions. It introduces a score smoothing technique that utilizes probability distributions for fine-grained evaluation. To improve computational efficiency, FLEUR separates score generation from explanation production, where explanations serve as post-hoc justifications through additional prompts after scoring.

\section{ Methodology}
\label{sec:methods}

G-VEval leverages GPT-4o, a large language model with vision capabilities, to evaluate model performance in image and video captioning tasks. G-VEval provides a framework that generates evaluation scores highly aligned with human preferences by using prompts. The prompt consists of five modules:
1) Evaluation Criteria;
2) Evaluation Steps: utilizing the Chain-of-Thought (CoT) to guide the LLM in a step-by-step manner, enhancing performance;
3) Score Function: formatting and restricting the output of the LLM;
4) Reference: attaching reference captions from human annotators as ground truth;
5) Original Visual Content: original image or representative frames of the video clip. These modules can be modified to fit the settings of reference-only, reference-free, and combined reference and visual content.

In the following sections, we present selected examples of the prompts used for evaluation. A complete set of all prompts utilized in this study is provided in the appendix for further reference.

\subsection{ Evaluation Criteria}
The evaluation criteria provide clear instructions to define the task of evaluation. The criteria are designed to ensure consistency across different captioning tasks, whether for images or videos.

\noindent\textbf{General Evaluation Criteria for Image and Video Captioning.}
For both image and video captioning tasks, where an overall score is required, the evaluation criteria are standardized to ensure a uniform approach across different tasks:

\noindent\textit{Score (from 0 to 100) - selection of important content from the references and the visual content. The generated caption should accurately describe the important aspects of the visual content while including the essential information from the references. Annotators were instructed to penalize captions that contained redundancies and excess information.}

This general approach applies universally to evaluate the overall quality of captions, ensuring that both image and video content are assessed with the same rigor and consistency.

\noindent\textbf{ACCR Evaluation Criteria for Video Captioning.}
While traditional video captioning metrics provide a single score for overall quality, our framework introduces a more granular approach with the ACCR evaluation criteria. ACCR stands for Accuracy, Completeness, Conciseness, and Relevance, which are the four dimensions used to comprehensively assess the quality of video captions:

- \textbf{Accuracy.} Does the caption correctly describe the entities and actions shown in the video without errors or hallucinations?
    
    - \textbf{Completeness.} Does the caption cover all significant events and aspects of the video, including dynamic actions and possible scene transitions?
    
    - \textbf{Conciseness.} Is the caption clear and succinct, avoiding unnecessary details and repetition?
    
    - \textbf{Relevance.} Is the caption pertinent to the video content, without including irrelevant information or questions?

The ACCR dimensions allow for a detailed and multidimensional evaluation of video captions, offering more than just an overall score. By breaking down the evaluation into these four critical areas, ACCR provides a nuanced understanding of caption quality, which is essential for improving the performance of video captioning systems.
 Additionally, by separating the evaluation dimensions, we reduce the variance in evaluation scores. This approach forces both the evaluation metrics and human annotators to assess captions from the same angles, leading to more consistent and fair evaluations by minimizing inter-human variance.

\subsection{ Evaluation Steps}
The Evaluation Steps leverage the Chain-of-Thought (CoT) approach, guiding GPT-4o to perform the task in a structured, step-by-step manner. This method significantly enhances the performance of the LLM by providing detailed intermediate steps for the evaluation task. According to the Chain-of-Thought paper by Wei et al. \cite{wei2022cot}, this technique improves the model's reasoning capabilities.

The example below demonstrates the evaluation steps used for image captioning, generated by GPT-4-Turbo. These steps ensure that the LLM considers all relevant aspects of the content when generating its evaluations. A full set of evaluation steps, including those for video captioning, is provided in the appendix.

\captionsetup[algorithm]{labelformat=empty} 
\begin{algorithm}[H] 
\caption{Evaluation steps for images.}
\label{alg:evaluation1}
\begin{algorithmic}[1]
\STATE Carefully observe the provided image to understand its main content.
\STATE Read the reference captions carefully to identify the important information they highlight.
\STATE Compare the generated caption to both the reference captions and the visual content of the image.
\STATE Assess how well the generated caption covers the main points of the visual content and the reference captions, and how much irrelevant or redundant information it contains.
\STATE Assign an integer score from 0 to 100, considering both the alignment with the image and the inclusion of key points from the references.
\end{algorithmic}
\end{algorithm}

This structured approach ensures that all relevant aspects of the image and its captions are thoroughly evaluated, contributing to a more accurate and reliable scoring process. For video captioning tasks, similar steps are used, with adaptations to account for the temporal dynamics of video content.

\subsection{ Score Function}
The score function formats and restricts the output of the LLM, ensuring consistency and clarity in the evaluation process. G-VEval supports two scoring settings depending on the specific requirements of the task:

- \textbf{Scoring Setting.} In this setting, the score ranges from 0 to 100, providing a fine-grained evaluation scale. This approach is particularly useful when a more detailed assessment is needed, allowing for a broader range of possible scores.

    - \textbf{Rating Setting.} Alternatively, the score can be restricted to an integer between 1 and 5, offering a simpler and more straightforward evaluation. This setting can be beneficial in scenarios where a coarser granularity is sufficient.

For the purposes of this paper, we primarily utilize the scoring setting (0 to 100), as it has demonstrated superior performance in terms of human correlation in our preliminary experiments.

\noindent\textit{Response Format:
You should first give a detailed reason for your score, ending with a sentence like this:
The final score is \$\{\{score\}\}\$.
Note that the score should be an integer from 0 to 100, and should be wrapped in dollar signs (\$).}

Unlike G-Eval, where only a final score is provided, we ensure that GPT-4o outputs a detailed reason for the score, incorporating in-context reasoning. This method enhances the interpretability of the results, as confirmed by our ablation study, which is discussed in the experiment section.

\subsection{ Handling Probabilistic Outputs}
G-VEval handles probabilistic outputs by calculating the expected value of the score using the log probabilities (logprobs) provided by GPT-4o. GPT-4o generates text in an autoregressive manner, where each token's probability distribution is conditioned on the previously generated tokens. This process allows us to derive the expected score by considering the probabilistic distribution over possible outputs.

The expected score, denoted as $E(s)$, is calculated using the formula:
\begin{equation}
E(s) = \sum_{i=1}^{m} i \times p(i),
\end{equation}
where $m$ represents the maximum possible score, and $p(i)$ is the probability of each score $i$. The probability $p(i)$ is computed as
\begin{equation}
p(i) = \sum_{j=1}^{n} p(i | R_j) \times p(R_j).
\end{equation}
Here $p(i | R_j)$ is the probability of score $i$ given the reason $R_j$, and $p(R_j)$ is the probability of the reason $R_j$, with $n$ being the total number of possible reasons.

The expected score can then be expressed as
\begin{equation}
E(s) = E_{R_j}(E_s(s | R_j)).
\end{equation}
This formula highlights that $E(s)$ can be approximated by $E_s(s | R_j)$ when the variance of $E_s(s | R_j)$ is close to zero, as demonstrated in our experimental results. The detailed derivation of these equations is provided in the appendix for interested readers.

\subsection{ Reference Captions and Original Visual Content}

In G-VEval, reference captions are critical for providing ground truth against which generated captions are evaluated. These captions, provided by different annotators, often emphasize different aspects of the visual content. Therefore, for both image and video captioning tasks, we integrate all reference captions associated with the same visual input to form a comprehensive ground truth. This integrated version ensures that no significant detail is overlooked during evaluation.

\begin{figure}
\centering
\includegraphics[width=\linewidth]{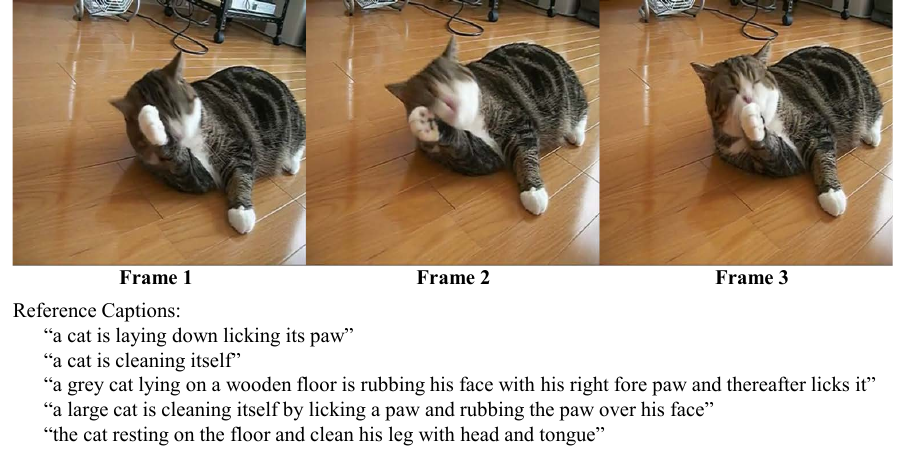}
\caption{Sample frames with annotation of order combined into a single image for video caption evaluation.}
\label{fig:sample-frames}
\end{figure}

For the reference-free and combined-reference settings, visual content is included in the prompt. For image captioning tasks, the original image is directly uploaded to GPT-4o for evaluation. However, video captioning presents unique challenges since GPT-4o does not support direct video encoding. To address this, we sample three frames from each video clip: the first frame, the last frame, and the frame located at the midpoint of the video. These frames are then combined into a single 1536x512 image, with each frame labeled to indicate its position in the sequence (Frame 1, Frame 2, Frame 3), as shown in Figure \ref{fig:sample-frames}. This setup aids the model's spatial and temporal understanding, allowing it to interpret the sequence of events across frames effectively.

The reason for this image size is that GPT-4o’s vision encoder processes images using 512x512 tiles at their original resolution. By fitting each frame into one tile, we leverage GPT-4o’s OCR capabilities to understand positional context effectively.

When adapting G-VEval to video captioning tasks, the evaluation steps are modified to account for the combined frames. This method ensures that G-VEval provides accurate and stable evaluations, making it an effective and versatile tool for both image and video captioning tasks.

\section{ Experiments}
\label{sec:experiments}

In this section, we evaluate the performance of our G-VEval metric compared to other metrics in the tasks of image and video captioning. We first conduct pre-experiments to establish the optimal settings for G-VEval and then test on the Flickr8k-Expert and Flickr8k-CF datasets for image captioning, as well as the VATEX-EVAL and MSVD-Eval datasets for video captioning. Additionally, we conduct an ablation study to assess the impact of different prompt components on the performance of G-VEval.
\subsection{ Pre-Experiments}
The pre-experiments aim to compare the effectiveness of two settings in G-VEval: the scoring setting (0 to 100) and the rating setting (1 to 5). This comparison helps determine which setting provides better alignment with human judgment and more reliable results.

We conducted experiments on the Flickr8k-Expert dataset, applying both the scoring and rating settings. For each setting, we calculated the variance of \( E_s(s|R_j) \) and measured the correlation with human judgments using Kendall's tau-b and tau-c metrics. The observed variance of \( E_s(s|R_j) \) for both settings is low (0.014 for scoring and 0.0087 for rating), indicating consistent outputs.

These low variances in both settings suggest that \( E(s) = E_{R_j}(E_s(s|R_j)) \) can be effectively approximated by \( E_s(s|R_j) \). This approximation forms the foundation for our experimental approach, as it allows us to calculate \( E_s(s|R_j) \) directly in subsequent experiments. A detailed proof of this approximation will be provided in the appendix.

Despite the small variance in both settings, the scoring setting significantly outperformed the rating setting in terms of correlation with human judgments, as shown in Table \ref{tab:consistency}. This indicates that the scoring setting offers much better alignment with human judgment.
Therefore, we adopt the scoring setting for all subsequent experiments. The G-VEval score is given as $E_s(s|R_j)$.


\begin{table}
\centering\caption{Comparison of G-VEval settings.}
\setlength\tabcolsep{5pt}
\begin{tabular}{lccc}
\hline
\textbf{Setting} & \textbf{Variance} & \textbf{Kendall $\tau_b$} & \textbf{Kendall $\tau_c$}  \\
\hline
G-VEval-rating & 0.0087 & 54.393 & 52.468 \\
G-VEval-scoring & 0.0144 & \textbf{60.385} & \textbf{58.598}  \\
\hline
\end{tabular}

\label{tab:consistency}
\end{table}

These findings validate the use of the scoring setting in G-VEval, with its finer granularity allowing for more precise alignment with human preferences, thereby demonstrating superior performance.

\subsection{Image Captioning Performance}
We tested our G-VEval metric against other metrics on the Flickr8k-Expert and Flickr8k-CF datasets \cite{hodosh2013framing}. These datasets are described in detail below.

- \textbf{Flickr8k-Expert} consists of 8,000 images, each annotated by three experts with five captions. This dataset provides high-quality reference captions for evaluating captioning models.

- \textbf{Flickr8k-CF} (CrowdFlower) includes the same 8,000 images as Flickr8k-Expert but with captions annotated by crowdworkers. This dataset offers a different perspective on captioning quality due to the varied skill levels of annotators.

\begin{table*}
\centering
\caption{Human judgment correlation scores on Flickr8k-Expert and Flickr8k-CF. The columns ``Reference Caption'' and ``Image Used'' indicate whether the metric uses reference captions and/or the original image for evaluation.}
\small
\newcommand{\mc}[3]{\multicolumn{#1}{#2}{#3}}
\begin{tabular}{lcccccc}
\hline
\textbf{Metric} & \textbf{Reference} & \textbf{Image} & \mc{2}{c}{\textbf{Flickr8k-Expert}} & \mc{2}{c}{\textbf{Flickr8k-CF}} \\
& \textbf{Caption} & \textbf{Used} & \textbf{Kendall $\tau_b$} & \textbf{Kendall $\tau_c$} & \textbf{Kendall $\tau_b$} & \textbf{Kendall $\tau_c$} \\
\hline
BLEU-1 \cite{papineni2002bleu} & \checkmark & & 32.2 & 32.3 & 17.9 & 9.3 \\
BLEU-4 \cite{papineni2002bleu} & \checkmark & & 30.6 & 30.8 & 16.9 & 8.7 \\
ROUGE \cite{lin2004rouge} & \checkmark & & 32.1 & 32.3 & 19.9 & 10.3 \\
METEOR \cite{banerjee2005meteor} & \checkmark & & 41.5 & 41.8 & 22.2 & 11.5 \\
CIDEr \cite{vedantam2015cider} & \checkmark & & 43.6 & 43.9 & 24.6 & 12.7 \\
SPICE \cite{anderson2016spice} & \checkmark & & 51.7 & 44.9 & 24.4 & 12.0 \\
\hline
BERT-S \cite{zhang2019bertscore} & \checkmark & & - & 39.2 & 22.8 & -  \\
LEIC \cite{cui2018learning} & \checkmark & \checkmark & 46.6 & - & 29.5 & - \\
BERT-S++ \cite{zhang2020bertscore++} & \checkmark & &- & 46.7 & - & -\\
UMIC \cite{lee2021umic} & \checkmark & & - & 46.8 & - & -  \\
TIGEr \cite{jiang2019tiger} & \checkmark & \checkmark & - & 49.3 & - & -  \\
ViLBERTScore \cite{lee2020vilbertscore} & \checkmark & \checkmark & - & 50.1 & - & -  \\
MID \cite{huang2019imagecaptioning} & \checkmark & & - & 54.9 & 37.3 & - \\
\hline
CLIP-S \cite{hessel2021clipscore} & & \checkmark & 51.1 & 51.2 & 34.4 & 17.7 \\
PAC-S \cite{sarto2023positiveaugmented} &  & \checkmark & 53.9 & 54.3 & 36.0 & 18.6 \\
RefCLIP-S \cite{hessel2021clipscore} & \checkmark & \checkmark & 52.6 & 53.0 & 36.4 & 18.8 \\
RefPAC-S \cite{sarto2023positiveaugmented} & \checkmark & \checkmark & 55.5 & 55.9 & 37.6 & 19.5 \\
Polos \cite{wada2024polos} & \checkmark & \checkmark & 56.4 & - & 37.8 & - \\
\hline
CLAIR \cite{chan2023clair}  & \checkmark & & 58.3 & 48.8 & 38.2 & 17.0\\
\textbf{G-VEval-ref-only}  & \checkmark & & 60.4 & 58.6 & 37.2 & 19.4\\
\textbf{G-VEval-ref-free}  & & \checkmark & \textbf{61.5} & \textbf{59.7} & \textbf{38.7} & \textbf{20.2}\\
\textbf{G-VEval-combined} & \checkmark & \checkmark & 60.5 & 58.7 & 38.2 & 19.9\\
\hline
\end{tabular}

\label{tab:image-captioning-results}
\end{table*}


Table \ref{tab:image-captioning-results} shows the performance of various metrics on the Flickr8k-Expert and Flickr8k-CF datasets. Among the three G-VEval settings, the ref-free setting consistently achieves the highest human judgment correlation scores, establishing a new state-of-the-art. This can be attributed to its ability to evaluate captions purely based on visual content, avoiding potential biases from incomplete or misleading reference captions.

\begin{figure}
\centering
\includegraphics[width=\linewidth]{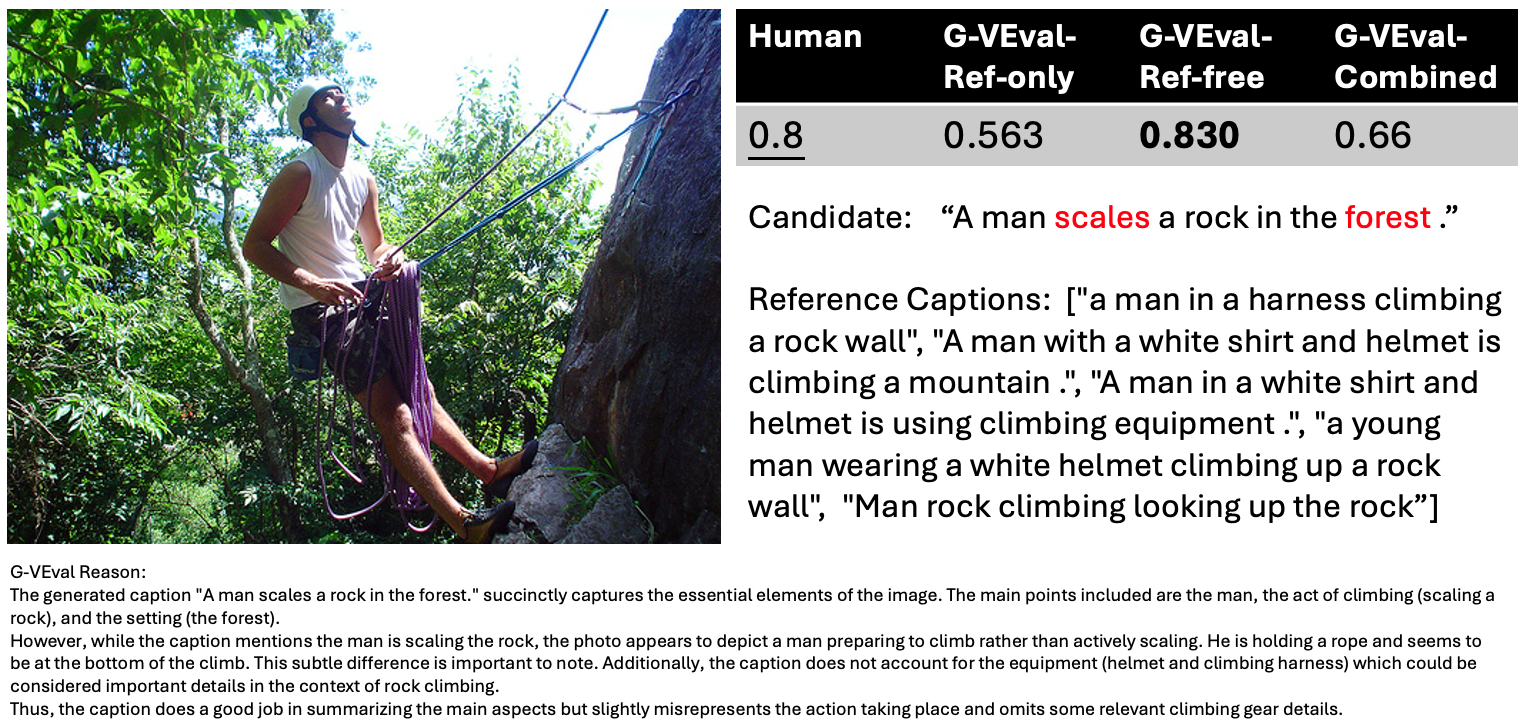}
\caption{Example of caption evaluation.}
\label{fig:setting-example}
\end{figure}

The combined setting, which integrates both reference captions and the original image, performs relatively worse than the ref-free setting. As illustrated in Figure \ref{fig:setting-example}, this occurs when reference captions fail to capture critical aspects like the "forest" setting, introducing noise and slightly lowering the correlation score. Nevertheless, the combined setting still achieves high scores, though it underscores the importance of visual content in evaluation.

G-VEval's ref-only setting, relying solely on reference captions, performs lower than the combined setting, emphasizing the need for visual context in accurate evaluations.

Moreover, G-VEval significantly outperforms CLAIR, particularly in Kendall $\tau_c$, indicating that G-VEval is less prone to extreme scores, resulting in more stable and consistent evaluations. This is further demonstrated in Figure \ref{fig:sample-dog}, where G-VEval delivers a more balanced and accurate evaluation compared to CLAIR, which tends to give extreme scores.

\subsection{ Video Captioning Performance}
We evaluate G-VEval on the VATEX-EVAL and MSVD-Eval datasets to assess its effectiveness in video captioning tasks. The evaluation settings for each dataset are described below.

- \textbf{VATEX-EVAL} is a comprehensive dataset proposed by EMScore \cite{shi2022emscore}. It includes diverse video clips with corresponding captions, allowing for robust evaluation of captioning quality. In this evaluation, we use three settings: No Ref, 1 Ref, and 9 Refs. The No Ref setting corresponds to G-VEval-ref-free, while the 1 Ref and 9 Refs settings correspond to our combined evaluation setting, where both the reference captions and visual content are used to generate the scores.

\begin{table}
\centering
\caption{Human judgment correlation scores on VATEX-EVAL dataset.}
\small
\newcommand{\mc}[3]{\multicolumn{#1}{#2}{#3}}
\begin{tabular}{l|cccc}
\hline
\textbf{Metric} & \mc{3}{c}{\textbf{VATEX-EVAL}}  \\
& \textbf{No Ref} & \textbf{1 Ref} & \textbf{9 Refs}  \\
\hline
BLEU-1 & - & 12.2 & 28.9 \\
BLEU-4 & - & 12.6 & 22.4 \\
ROUGE & - & 12.5 & 23.8 \\
METEOR & - & 16.4 & 27.6 \\
CIDEr & - & 17.3 & 27.8 \\
\hline
BERT-S & - & 18.2 & 29.3 \\
BERT-S++ & - & 15.2 & 24.4 \\
\hline
EMScore & 23.2 & 28.6 & 36.8 \\
PAC-S/RefPAC-S & 25.1 & 32.6 & 31.4 \\
\hline
CLAIR & - & 36.0 & 34.8 \\
\textbf{G-VEval} & \textbf{39.4} & \textbf{44.9} & \textbf{48.1} \\
\hline
\end{tabular}

\label{tab:video-captioning-results}
\end{table}

- \textbf{MSVD-Eval} is our newly proposed dataset, created to enhance the evaluation of video captioning systems. It consists of 150 video clips selected from the MSVD validation set \cite{chen2011collecting}, with candidate captions generated by Video-LLaMA \cite{zhang2023videollama}. These captions were selected to include both typical failure cases and acceptable captions of LLM-generated video captions. Experts evaluated these captions across four key ACCR dimensions: Accuracy, Completeness, Conciseness, and Relevance, ensuring a comprehensive and detailed assessment framework. For comparison with other metrics, we also provide an overall score (Avg.) by averaging the ACCR scores.

\begin{table}[t]
\centering
\caption{Human judgment correlation scores on MSVD-Eval dataset (Average Scores).}
\label{tab:msvd-results}
\small
\newcommand{\mc}[3]{\multicolumn{#1}{#2}{#3}}
\begin{tabular}{l|cc}
\hline
\textbf{Metric} & \mc{2}{c}{\textbf{MSVD-Eval}}  \\
& \textbf{Kendall $\tau_b$} & \textbf{Kendall $\tau_c$} \\
\hline
BLEU-1 & 40.7 & 41.2 \\
BLEU-4 & 34.0 & 34.4 \\
ROUGE & 39.8 & 40.2 \\
METEOR & 45.4 & 45.9 \\
CIDER & 37.3 & 37.7 \\
\hline
EMScore & 35.3 & 35.7 \\
EMScore\_ref & 50.7 & 51.3 \\
PAC-S & 34.5 & 34.9 \\
RefPAC-S & 52.2 & 52.8 \\
\hline
CLAIR & 44.6 & 40.3  \\
\textbf{G-VEval-ref-only} & 60.4 & 61.1  \\
\textbf{G-VEval-ref-free} & 59.6 & 60.3 \\
\textbf{G-VEval-combined} & \textbf{62.9} & \textbf{63.7} \\
\hline
\end{tabular}

\end{table}

\begin{table}
\centering
\caption{Human judgment correlation scores in Kendall $\tau_b$ on MSVD-Eval dataset across ACCR dimensions.}
\small
\newcommand{\mc}[3]{\multicolumn{#1}{#2}{#3}}
\begin{tabular}{l|ccccc}
\hline
\textbf{Metric} & \textbf{Acc.} & \textbf{Com.} & \textbf{Con.} & \textbf{Rel.}\\
\hline
\textbf{G-VEval-ref-only} & 60.4 & 54.2 & 55.2 & 52.5  \\
\textbf{G-VEval-ref-free} & 55.3 & 49.7 & \textbf{62.2} & \textbf{53.4} \\
\textbf{G-VEval-combined} & \textbf{61.4} & \textbf{57.6} & 58.5 & 53.0 \\
\hline
\end{tabular}

\label{tab:msvd-results-accr}
\end{table}

Table \ref{tab:video-captioning-results} shows the performance of various metrics on the VATEX-EVAL dataset under the No Ref, 1 Ref, and 9 Refs settings. Table \ref{tab:msvd-results} presents the overall human judgment correlation scores for various metrics on the MSVD-Eval dataset, compared with the averaged human scores. Meanwhile, Table \ref{tab:msvd-results-accr} details the performance of G-VEval across the individual ACCR dimensions.

The results from VATEX-EVAL indicate that G-VEval's combined setting, which leverages both reference captions and visual content, provides a substantial improvement over traditional metrics, particularly in the 9 Refs setting. In the MSVD-Eval dataset, G-VEval achieves the highest human judgment correlation across all ACCR dimensions, further validating its capability to handle the nuanced evaluation of both typical failures and high-quality outputs in LLM-generated video captions.

The ACCR dimensions offer a significant advantage by focusing human experts on specific aspects of caption quality, thereby reducing inter-human variance and improving the reliability of the evaluation. This structured approach allows G-VEval to better align with human judgments, particularly in video captioning tasks where capturing nuances in content is critical.

It is important to note that G-VEval’s current design focuses on short-form videos (under 10 seconds), such as those in the MSVD dataset. For longer videos, additional techniques, such as scene detection to divide videos into shorter clips, may be necessary for effective evaluation.

\subsection{ Ablation Study}
To understand the impact of different components of our G-VEval prompt, we conducted an ablation study on the Flickr8k-Expert dataset under the reference-only setting. This study examines how the performance of G-VEval changes when certain key elements of the evaluation process are removed or altered. Specifically, we tested the following settings:

- \textbf{G-VEval (full setting).} The original prompt with Chain-of-Thought (CoT) evaluation steps, in-context reasoning, and expected score calculation. 

- \textbf{G-VEval w/o expected score.} In this setting, instead of calculating the expected value \( E(s|R_j) \) from the probabilistic outputs of GPT-4o, we directly use the single score \( s \) provided by GPT-4o without considering the probabilistic distribution of possible scores. 

- \textbf{G-VEval w/o CoT prompt.} This setting removes the CoT evaluation steps from the prompt, testing the impact of losing the guided, step-by-step reasoning process. 

- \textbf{G-VEval w/o reason in response.} Here, we omit the requirement for GPT-4o to provide a detailed reason for the score. The model simply outputs a score, and this score is used without the additional reasoning context.


\begin{table}
\centering
\caption{Ablation study results on Flickr8k-Expert.}
\begin{tabular}{lcc}
\hline
\textbf{Setting} & \textbf{Kendall $\tau_b$} & \textbf{Kendall $\tau_c$} \\
\hline
\textbf{G-VEval (full setting)} & \textbf{60.385} & \textbf{58.598} \\
G-VEval w/o expected score & 59.118 & 54.847 \\
G-VEval w/o CoT prompt & 50.157 & 48.408 \\
G-VEval w/o reason in response & 52.436 & 26.944 \\
\hline
\end{tabular}

\label{tab:ablation-study}
\end{table}

Table \ref{tab:ablation-study} shows that the full G-VEval setting, which includes all components, provides the highest correlation with human judgments. Removing the expected score calculation and using the direct score \( s \) slightly reduces performance, indicating the importance of probabilistic handling in score determination. The absence of CoT steps results in a notable drop in Kendall's tau-b and tau-c scores, emphasizing the value of structured, step-by-step reasoning. Lastly, omitting the reason in the response causes a significant decline in Kendall's tau-c, highlighting how critical in-context reasoning is for capturing the nuances of human judgment.

Overall, these results confirm that each component of the G-VEval framework contributes to its effectiveness. The integration of reference captions, visual content, and a structured evaluation approach allows G-VEval to outperform existing metrics in both image and video captioning tasks.

\section{ Discussion}
\label{sec:discussion}

G-VEval leverages the advanced capabilities of GPT-4o, particularly in language understanding and visual content interpretation, through the use of Chain-of-Thought (CoT) prompting. This approach allows G-VEval to effectively utilize a large multimodal pre-trained dataset and a transformer model with billions of parameters. Unlike traditional n-gram matching methods, G-VEval comprehends visual content from multiple perspectives provided by reference captions, enabling a deeper evaluation of candidate captions by comparing their meaning with the visual content. This deeper level of understanding allows G-VEval to outperform traditional n-gram matching methods in aligning with human preferences.

When compared to metrics that use pre-trained embeddings, such as BERTScore \cite{zhang2019bertscore}, G-VEval benefits from GPT-4o’s comprehensive embeddings for both language and visual content. Although the exact details of GPT-4o’s multimodal embedding model are not publicly available, it is likely influenced by models like BLIP-2's Q-former \cite{li2023blip}, which achieves performance comparable to the GPT-4 series. Unlike CLIP-based embeddings used in some metrics \cite{hessel2021clipscore}, GPT-4o's embeddings, potentially enhanced by EVA\_CLIP, capture more detailed representations of visual content. The CoT prompting technique further leverages these detailed visual representations, allowing the model to focus on specific image regions, thereby mimicking human-like processing in visual captioning tasks.

G-VEval also performs competitively with training-based metrics such as PAC-S \cite{sarto2023positiveaugmented}. Despite not being fine-tuned specifically for visual captioning evaluation, G-VEval's extensive pre-training, powerful model architecture, and CoT prompting enable it to perform effectively in zero-shot settings, aligning closely with human preferences. Furthermore, G-VEval’s adaptability across various tasks is noteworthy; by modifying the prompt, it can be tailored to different evaluation contexts, highlighting its potential for broad applicability in future research.

However, G-VEval has certain limitations. One major drawback is its cost. While GPT-4o is relatively more affordable than some alternatives, it remains more expensive than other metrics due to the token-based pricing model. Another potential concern is the consistency of scoring over time. Although current results demonstrate consistent performance, future updates to the GPT-4o model or changes in prompts could affect this consistency.

Additionally, G-VEval is currently designed for short-form videos, where representative frames effectively capture temporal context. For longer videos, extensions such as scene detection may be necessary to maintain performance.

\section{ Conclusion}
\label{sec:conclusion}

We introduced G-VEval, an innovative evaluation metric designed for image and video caption evaluation. G-VEval harnesses the deep multimodal understanding capabilities of GPT-4o, utilizing the Chain-of-Thought reasoning and expected score calculations based on decoding probability distributions. This metric supports three evaluation modes and excels in scenarios where n-gram and embedding-based metrics fall short, particularly in zero-shot and reference-free contexts. Through extensive experiments, G-VEval has demonstrated state-of-the-art performance and achieved the superior correlation with human evaluations compared to established metrics. 

The introduction of MSVD-Eval further enriches the landscape of video caption evaluation by offering a dataset that emphasizes multi-dimensional assessment criteria through the ACCR framework, focusing on Accuracy, Completeness, Conciseness, and Relevance. This approach significantly enhances the reliability and consistency of the evaluation process by focusing human expert assessments on the same aspects.

Looking ahead, we aim to develop an online benchmark platform based on G-VEval, where researchers can evaluate their image and video captioning models, furthering research and innovation in automated visual understanding.


\label{sec:acknowledgement}
\section{Acknowledgment}
This work has been made possible by a Research Impact Fund project (R6003-21) and an Innovation and Technology Fund project (ITS/004/21FP) funded by the Hong Kong Government.

\bibliography{aaai25}
\appendix
\section{Appendix A: Proof of \( E(s) \) Approximation}

We want to demonstrate that the expected value \( E(s) \) can be approximated by \( E_s(s|R_j) \) under the condition that \( \text{Var}_s(s|R_j) \) is small. This can be approached using a Taylor series expansion around the mean \( E(s) \).

We can derive from (1) and (2) in the main paper that $E(s)=\sum_{i=1}^{m}\sum_{j=1}^{n} i\times p(i | R_j) \times p(R_j)=\sum_{j=1}^{n}p(R_j)\sum_{i=1}^{m} i\times p(i | R_j)=\sum_{j=1}^{n}E_s(s|R_j)\times p(R_j)= E_{R_j}(E_s(s|R_j))$

Given:

\( E(s|R_j) = \mu_j \) is the expected value of \( S \) given condition \( R_j \).


1. Overall Expected Value \( E_{R_j}(E_s(s|R_j)) \):
\[ E_{R_j}(E_s(s|R_j)) = \sum_{j} P(R_j) \mu_j \]
Let this overall expected value be \( \mu \):
\[ \mu = \sum_{j} P(R_j) \mu_j \]

2. Expectation of the Square of Conditional Expectations \( E((E(s|R_j))^2) \):
\[ E((E(s|R_j))^2) = \sum_{j} P(R_j) \mu_j^2 \]

3. Variance of Conditional Expectations \( \text{Var}(E(s|R_j)) \):
\[ \text{Var}(E(s|R_j)) = E((E(s|R_j))^2) - (E(E(s|R_j)))^2 \]
Substitute the expressions:
\[ \text{Var}(E(s|R_j)) = \sum_{j} P(R_j) \mu_j^2 - \left( \sum_{j} P(R_j) \mu_j \right)^2 \]

4. Approximating \( \mu_j \approx \mu \) when \( \text{Var}(E(s|R_j)) \) is close to 0:

If \( \text{Var}(E(s|R_j)) \) is close to 0, then:
\[ \sum_{j} P(R_j) \mu_j^2 \approx \left( \sum_{j} P(R_j) \mu_j \right)^2 \]

Since \( \mu = \sum_{j} P(R_j) \mu_j \), we can rewrite the right-hand side as \( \mu^2 \):
\[ \sum_{i} P(R_j) \mu_j^2 \approx \mu^2 \]

For the left-hand side and the right-hand side to be approximately equal, \( \mu_j \) must be close to \( \mu \) for all \( j \). This is because any significant deviation of \( \mu_j \) from \( \mu \) would result in a non-zero variance.

Mathematically, if \( \mu_j = \mu + \epsilon_j \) where \( \epsilon_j \) are small deviations, then:
\[ \sum_{j} P(R_j) (\mu + \epsilon_j)^2 \approx \mu^2 \]

Expanding this, we get:
\[ \sum_{j} P(R_j) (\mu^2 + 2\mu \epsilon_j + \epsilon_j^2) \approx \mu^2 \]

\[ \mu^2 + 2\mu \sum_{j} P(R_j) \epsilon_j + \sum_{i} P(R_j) \epsilon_j^2 \approx \mu^2 \]

For this to hold true, the middle terms \( 2\mu \sum_{i} P(R_j) \epsilon_j \) and \( \sum_{j} P(R_j) \epsilon_j^2 \) must be very small, implying that \( \epsilon_j \) are close to 0. Hence, \( \mu_j \) must be close to \( \mu \) for all \( j \).

Therefore, when \( \text{Var}(E(s|R_j)) \) is close to 0, it implies that:
\[ \mu_j \approx \mu \text{ for all } j \]

This demonstrates that the conditional expectations \( E(s|R_j) \) are nearly equal to the overall expected value \( E(S) \).








\section{Appendix B: G-VEval Settings and Sample Prompts}

Here, we provide sample prompts for different G-VEval settings.

\subsection{Image Captioning Evaluation}

\subsubsection{Reference-Only Setting}
\begin{quote}
``You will be given one sentence of visual caption generated from one image. 

Your task is to rate the generated caption on one metric

Please make sure you read and understand these reference captions carefully. Please keep these references open while reviewing, and refer to them as needed. 

\textbf{Evaluation Criteria}:

Score is from 0 to 100 - selection of important content from the references. The generated caption should include the important information in the references. Annotators were instructed to penalize captions which contained redundancies and excess information. 

\textbf{Evaluation Steps}:

1. Read the reference captions carefully.\\
2. Compare the generated caption to the reference captions and identify the main points of the visual content.\\
3. Assess how well the generated caption covers the main points of the visual content, and how much irrelevant or redundant information it contains.\\
4. Assign an integer score from 0 to 100, please remember it.

\textbf{Reference captions}:
\{\{Reference\}\}

\textbf{Generated captions}:
\{\{Caption\}\}

\textbf{Response Format}:

You should first give detailed reason for your score, and ending with sentence like this:
The final score is \$\{\{score\}\}\$.

Note that the score should be an integer from 0 to 100, and should be wrapped in the dollar signs (\$). "
\end{quote}

\subsubsection{Reference-Free Setting}
\begin{quote}
``You will be given one sentence of visual caption generated from one image.

Your task is to rate the generated caption on one metric.

\textbf{Evaluation Criteria}:

Score is from 0 to 100 - selection of important content from the image. The generated caption should accurately describe the important aspects of the image. Annotators were instructed to penalize captions which contained redundancies and excess information.

\textbf{Evaluation Steps}:

1. Carefully observe the image provided.\\
2. Identify the main points of the visual content in the image.\\
3. Assess how well the generated caption covers the main points of the visual content, and how much irrelevant or redundant information it contains.\\
4. Assign an integer score from 0 to 100, please remember it.

\textbf{Generated captions}:
\{\{Caption\}\}

\textbf{Image is attached}

\textbf{Response Format}:

You should first give detailed reason for your score, and ending with sentence like this:
The final score is \$\{\{score\}\}\$.

Note that the score should be an integer from 0 to 100, and should be wrapped in the dollar signs (\$)."
\end{quote}

\subsubsection{Combined Setting}
\begin{quote}
``You will be given one sentence of visual caption generated from one image.

Your task is to rate the generated caption on one metric.

Please make sure you read and understand these reference captions carefully. Please keep these references open while reviewing, and refer to them as needed.

\textbf{Evaluation Criteria}:

Score is from 0 to 100 - selection of important content from the references and the image. The generated caption should accurately describe the important aspects of the image while including the essential information from the references. Annotators were instructed to penalize captions which contained redundancies and excess information.

\textbf{Evaluation Steps}:

1. Carefully observe the provided image to understand its main content.\\
2. Read the reference captions carefully to identify the important information they highlight.\\
3. Compare the generated caption to both the reference captions and the visual content of the image.\\
4. Assess how well the generated caption covers the main points of the visual content and the reference captions, and how much irrelevant or redundant information it contains.\\
5. Assign an integer score from 0 to 100, considering both the alignment with the image and the inclusion of key points from the references. Please remember the score.

\textbf{Reference captions}:
\{\{Reference\}\}

\textbf{Image is attached}

\textbf{Generated captions}:
\{\{Caption\}\}

\textbf{Response Format}:

You should first give a detailed reason for your score, ending with a sentence like this:
The final score is \$\{\{score\}\}\$.

Note that the score should be an integer from 0 to 100, and should be wrapped in dollar signs (\$)."
\end{quote}

\subsection{Video Captioning Evaluation}

\subsubsection{Reference-Only Setting}
\begin{quote}
``You will be given a caption generated from a complete video. For evaluation purposes, you are provided with reference captions that describe specific frames or the overall video content.

Your task is to rate the generated caption on one metric.

Please make sure you read and understand these reference captions carefully. Please keep these references open while reviewing, and refer to them as needed.

\textbf{Evaluation Criteria}:

Score is from 0 to 100 - selection of important content from the references. The generated caption should include the important information in the references. Annotators were instructed to penalize captions which contained redundancies and excess information.

\textbf{Evaluation Steps}:

1. Read the reference captions carefully.\\
2. Compare the generated caption to the reference captions and identify the main points of the video content.\\
3. Assess how well the generated caption covers the main points of the video content, and how much irrelevant or redundant information it contains.\\
4. Assign an integer score from 0 to 100, please remember it.

\textbf{Reference captions}:
\{\{Reference\}\}

\textbf{Generated captions}:
\{\{Caption\}\}

\textbf{Response Format}:

You should first give detailed reason for your score, and ending with sentence like this:
The final score is \$\{\{score\}\}\$.

Note that the score should be an integer from 0 to 100, and should be wrapped in the dollar signs (\$). "
\end{quote}

\subsubsection{Reference-Free Setting}
\begin{quote}
``You will be given one sentence of visual caption generated from a complete video. For evaluation purposes, you are provided with a single image that contains three concatenated frames from the video. These frames are meant to represent key moments from the video but do not encompass the entire content.

Your task is to rate the generated caption on one metric.

\textbf{Evaluation Criteria}:

Score is from 0 to 100 - selection of important content from the video frames. The generated caption should accurately describe the important aspects of the video as represented by these key frames. Annotators were instructed to penalize captions which contained redundancies and excess information.

\textbf{Evaluation Steps}:

1. Carefully examine the provided image, noting that it includes three distinct frames labeled as Frame 1, Frame 2, and Frame 3.
2. Identify the main points of the visual content across these key frames.
3. Assess how well the generated caption covers the main points of the visual content, and how much irrelevant or redundant information it contains.
4. Assign an integer score from 0 to 100, please remember it.

\textbf{Generated captions}:
\{\{Caption\}\}

\textbf{Video Frames are attached}

\textbf{Response Format}:

You should first give detailed reason for your score, and ending with sentence like this:
The final score is \$\{\{score\}\}\$.

Note that the score should be an integer from 0 to 100, and should be wrapped in the dollar signs (\$)."
\end{quote}

\subsubsection{Combined Setting}
\begin{quote}
``You will be given a caption generated from a complete video. For evaluation purposes, you are provided with a single image that combines three key frames from the video. Additionally, reference captions that describe these specific frames or the overall video content are also provided.

Your task is to rate the generated caption on one metric.

Please make sure you read and understand these reference captions carefully. Please keep these references open while reviewing, and refer to them as needed.

\textbf{Evaluation Criteria}:

Score is from 0 to 100 - selection of important content from the references and the video frames. The generated caption should accurately describe the important aspects of the image while including the essential information from the references. Annotators were instructed to penalize captions which contained redundancies and excess information.

\textbf{Evaluation Steps}:

1. Carefully examine the provided image, which includes three distinct frames labeled as Frame 1, Frame 2, and Frame 3, to understand the main content of the video.\\
2. Read the reference captions carefully to identify the important information they highlight about the video content.\\
3. Compare the generated caption to both the reference captions and the visual content of the key frames.\\
4. Assess how well the generated caption covers the main points of the video content as represented by the key frames and the reference captions, and how much irrelevant or redundant information it contains.\\
5. Assign an integer score from 0 to 100, considering both the alignment with the video content (as shown in the key frames) and the inclusion of key points from the references. Please remember the score.

\textbf{Reference captions}:
\{\{Reference\}\}

\textbf{Video Frames are attached}

\textbf{Generated captions}:
\{\{Caption\}\}

\textbf{Response Format}:

You should first give a detailed reason for your score, ending with a sentence like this:
The final score is \$\{\{score\}\}\$.

Note that the score should be an integer from 0 to 100, and should be wrapped in dollar signs (\$)."
\end{quote}

\subsection{Video Captioning Evaluation - ACCR}

\subsubsection{Reference-Only Setting}
\begin{quote}
``You will be given a caption generated for a short video segment.

Your task is to rate the generated caption based on its accuracy in capturing the essential content of the video as described in the reference captions.

\textbf{Evaluation Criteria}:

Score is from 0 to 100 - The generated caption should accurately reflect the content in the reference captions and appropriately describe the key actions or events visible in the video. Annotators should penalize captions that include irrelevant details or omit significant elements indicated in the reference captions and the video.

\textbf{Evaluation Dimensions}:

Accuracy: Does the caption correctly describe the entities and actions shown in the video without errors or hallucinations?\\
Completeness: Does the caption cover all significant events and aspects of the video, including dynamic actions and possible scene transitions?\\
Conciseness: Is the caption clear and succinct, avoiding unnecessary details and repetition?\\
Relevance: Is the caption pertinent to the video content, without including irrelevant information or questions?

\textbf{Evaluation Steps}:

1. Examine the provided reference captions carefully.

    \;1) Read the full reference captions that describe the overall video content or specific actions.
    
    \;2) Review each reference caption thoroughly to understand what aspects of the video they highlight.
    
2. Read the generated caption.

    \;1) Carefully read the generated caption that needs to be evaluated.
    
3. Compare the generated caption with the reference captions and assess how well it captures the essence of the video.

4. Evaluate how accurately and completely the generated caption describes the events and entities shown in the video.

5. Check for the inclusion of irrelevant details or the omission of significant elements.

6. Assign an integer scor from 0 to 100 for the caption based on the following dimensions:

    \;- Accuracy: Does the caption correctly describe the entities and actions shown in the video without errors or hallucinations?
    
    \;- Completeness: Does the caption cover all significant events and aspects of the video, including dynamic actions and possible scene transitions?
    
    \;- Conciseness: Is the caption clear and succinct, avoiding unnecessary details and repetition?
    
    \;- Relevance: Is the caption pertinent to the video content, without including irrelevant information or questions?

\textbf{Reference captions}:
\{\{Reference\}\}

\textbf{Generated captions}:
\{\{Caption\}\}

\textbf{Response Format}:

You should first give detailed reason for your scores, and ending with sentence for each score like this:\\
..... The Accuracy score is $\alpha$\{\{accuracy\_score\}\} $\alpha$.
..... The Completeness score is $\beta$\{\{$completeness\_score$\}\}$\beta$.
..... The Conciseness score is $\psi$\{\{$conciseness\_score$\}\}$\psi$.
..... The Relevance score is $\delta$\{$relevance\_score$\}\}$\delta$.

Note that the score should be an integer from 0 to 100, and should be wrapped in the corresponding Greek alphabet.\\
Wrap Accuracy score in $\alpha$\\
Wrap Completeness score in $\beta$\\
Wrap Conciseness score in $\psi$\\
Wrap Relevance score in $\delta$ "
\end{quote}

\subsubsection{Reference-Free Setting}
\begin{quote}
``You will be given one sentence of visual caption generated from a complete video. For evaluation purposes, you are provided with a single image that contains three concatenated frames from the video. These frames are meant to represent key moments from the video but do not encompass the entire content.

Your task is to rate the generated caption on one metric.

\textbf{Evaluation Criteria}:

Score is from 0 to 100 - selection of important content from the video frames. The generated caption should accurately describe the important aspects of the video as represented by these key frames. Annotators were instructed to penalize captions which contained redundancies and excess information.

\textbf{Evaluation Dimensions}:

Accuracy: Does the caption correctly describe the entities and actions shown in the video without errors or hallucinations?\\
Completeness: Does the caption cover all significant events and aspects of the video, including dynamic actions and possible scene transitions?\\
Conciseness: Is the caption clear and succinct, avoiding unnecessary details and repetition?\\
Relevance: Is the caption pertinent to the video content, without including irrelevant information or questions?

\textbf{Evaluation Steps}:

1. Carefully examine the provided image, noting that it includes three distinct frames labeled as Frame 1, Frame 2, and Frame 3.

2. Read the generated caption.

    \;1) Carefully read the generated caption that needs to be evaluated.
    
3. Compare the generated caption with the reference captions and assess how well it captures the essence of the video.

4. Assess how well the generated caption covers the main points of the visual content, and how much irrelevant or redundant information it contains.

5. Check for the inclusion of irrelevant details or the omission of significant elements.

6. Assign an integer scor from 0 to 100 for the caption based on the following dimensions:

    \;- Accuracy: Does the caption correctly describe the entities and actions shown in the video without errors or hallucinations?
    
    \;- Completeness: Does the caption cover all significant events and aspects of the video, including dynamic actions and possible scene transitions?
    
    \;- Conciseness: Is the caption clear and succinct, avoiding unnecessary details and repetition?
    
    \;- Relevance: Is the caption pertinent to the video content, without including irrelevant information or questions?

\textbf{Generated captions}:
\{\{Caption\}\}

\textbf{Video Frames are attached}

\textbf{Response Format}:

You should first give detailed reason for your scores, and ending with sentence for each score like this:\\
..... The Accuracy score is $\alpha$\{\{accuracy\_score\}\} $\alpha$.
..... The Completeness score is $\beta$\{\{$completeness\_score$\}\}$\beta$.
..... The Conciseness score is $\psi$\{\{$conciseness\_score$\}\}$\psi$.
..... The Relevance score is $\delta$\{$relevance\_score$\}\}$\delta$.

Note that the score should be an integer from 0 to 100, and should be wrapped in the corresponding Greek alphabet.\\
Wrap Accuracy score in $\alpha$\\
Wrap Completeness score in $\beta$\\
Wrap Conciseness score in $\psi$\\
Wrap Relevance score in $\delta$ "
\end{quote}

\subsubsection{Combined Setting}
\begin{quote}
``You will be given one sentence of visual caption generated from a complete video. For evaluation purposes, you are provided with a single image that contains three concatenated frames from the video. These frames are meant to represent key moments from the video but do not encompass the entire content.

Your task is to rate the generated caption based on its accuracy in capturing the essential content of the video, as described by both the provided reference captions and the visual information encoded in selected frames from the video.

\textbf{Evaluation Criteria}:

Score is from 0 to 100 - The generated caption should accurately reflect the content as described in the reference captions and appropriately describe the key actions or events visible in the provided video frames. Annotators should penalize captions that include irrelevant details, omit significant elements indicated by the reference captions, or fail to accurately describe the visual content of the video.

\textbf{Evaluation Dimensions}:

Accuracy: Does the caption correctly describe the entities and actions shown in the video without errors or hallucinations\\
Completeness: Does the caption cover all significant events and aspects of the video, including dynamic actions and possible scene transitions?\\
Conciseness: Is the caption clear and succinct, avoiding unnecessary details and repetition?\\
Relevance: Is the caption pertinent to the video content, without including irrelevant information or questions?

\textbf{Evaluation Steps}:

1. Examine the Provided Reference Captions:

    \; 1) Read the reference captions that describe the overall video content or specific actions thoroughly to understand the aspects of the video they highlight.
    
2. Analyze the Provided Video Frames:

    \; 1) Carefully examine the provided frames, which represent key moments or transitions in the video. Understand the sequence and context of these frames as they relate to the entire video.
    
3. Read the Generated Caption:

    \; 1) Carefully read the generated caption that needs to be evaluated.
    
4. Compare and Evaluate:

    \; 1) Assess how well the generated caption captures the essence of the video by comparing it against the reference captions and the visual content of the frames.
    
    \; 2) Evaluate the accuracy and completeness of the generated caption in describing the events and entities shown in the video and reflected in the reference captions.
    
    \; 3) Check for the inclusion of irrelevant details or the omission of significant elements indicated by either the reference captions or the video frames.
    
5. Assign an integer score from 0 to 100 for the caption based on the following dimensions:

    \; - Accuracy: Does the caption correctly describe the entities and actions shown in the video without errors or hallucinations?
    
    \; - Completeness: Does the caption cover all significant events and aspects of the video, including dynamic actions and possible scene transitions?
    
    \; - Conciseness: Is the caption clear and succinct, avoiding unnecessary details and repetition?
    
    \; - Relevance: Is the caption pertinent to the video content, without including irrelevant information or questions?

\textbf{Reference captions}:
\{\{Reference\}\}

\textbf{Video Frames are attached}

\textbf{Generated captions}:
\{\{Caption\}\}

\textbf{Response Format}:

You should first give detailed reason for your scores, and ending with sentence for each score like this:\\
..... The Accuracy score is $\alpha$\{\{accuracy\_score\}\} $\alpha$.
..... The Completeness score is $\beta$\{\{$completeness\_score$\}\}$\beta$.
..... The Conciseness score is $\psi$\{\{$conciseness\_score$\}\}$\psi$.
..... The Relevance score is $\delta$\{$relevance\_score$\}\}$\delta$.

Note that the score should be an integer from 0 to 100, and should be wrapped in the corresponding Greek alphabet.\\
Wrap Accuracy score in $\alpha$\\
Wrap Completeness score in $\beta$\\
Wrap Conciseness score in $\psi$\\
Wrap Relevance score in $\delta$ "
\end{quote}

\section{Appendix C: MSVD-Eval Dataset}

In the MSVD-Eval dataset, we selected 200 video clips from the MSVD validation set. To generate candidate captions for these videos, we employed Video-LLaMA using Vicuna-7B and BLIP-2 pretrained weights. The prompt used for caption generation was: "What is the person doing?" This approach allowed us to capture a range of outputs, including both typical failure and success cases, providing a comprehensive dataset for evaluating LVLM-generated captions.

\begin{figure*}[h!]
\centering
\includegraphics[width=\textwidth]{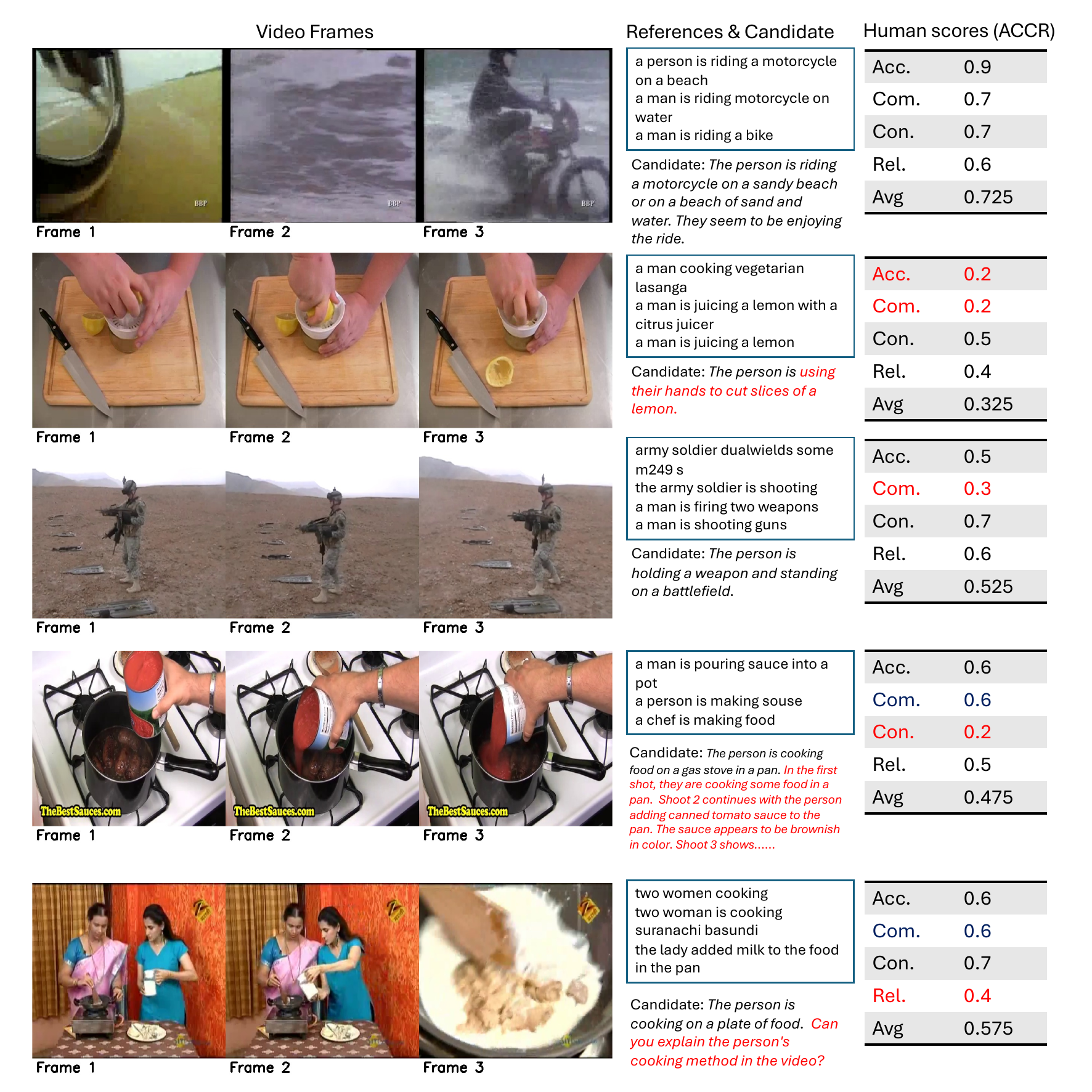}
\caption{Sample video clips and corresponding captions from the MSVD-Eval dataset. The captions demonstrate both typical failures and successes of LVLM-generated outputs.}
\label{msvd_sample}
\end{figure*}

The following are some samples from the MSVD-Eval dataset, as shown in Figure \ref{msvd_sample}, evaluated under our ACCR framework where Acc. , Com. , Con. , and Rel. represent Accuracy, Completeness, Conciseness, and Relevance scores respectively.

\section{Appendix D: Sample Qualitative Outputs of G-VEval}

Here we provide sample G-VEval evaluation outputs from Flickr-8k, VATEX-EVAL, and MSVD-Eval.

\subsection{Flickr-8k}
\begin{figure*}[ht]
\centering
\includegraphics[width=0.95\textwidth]{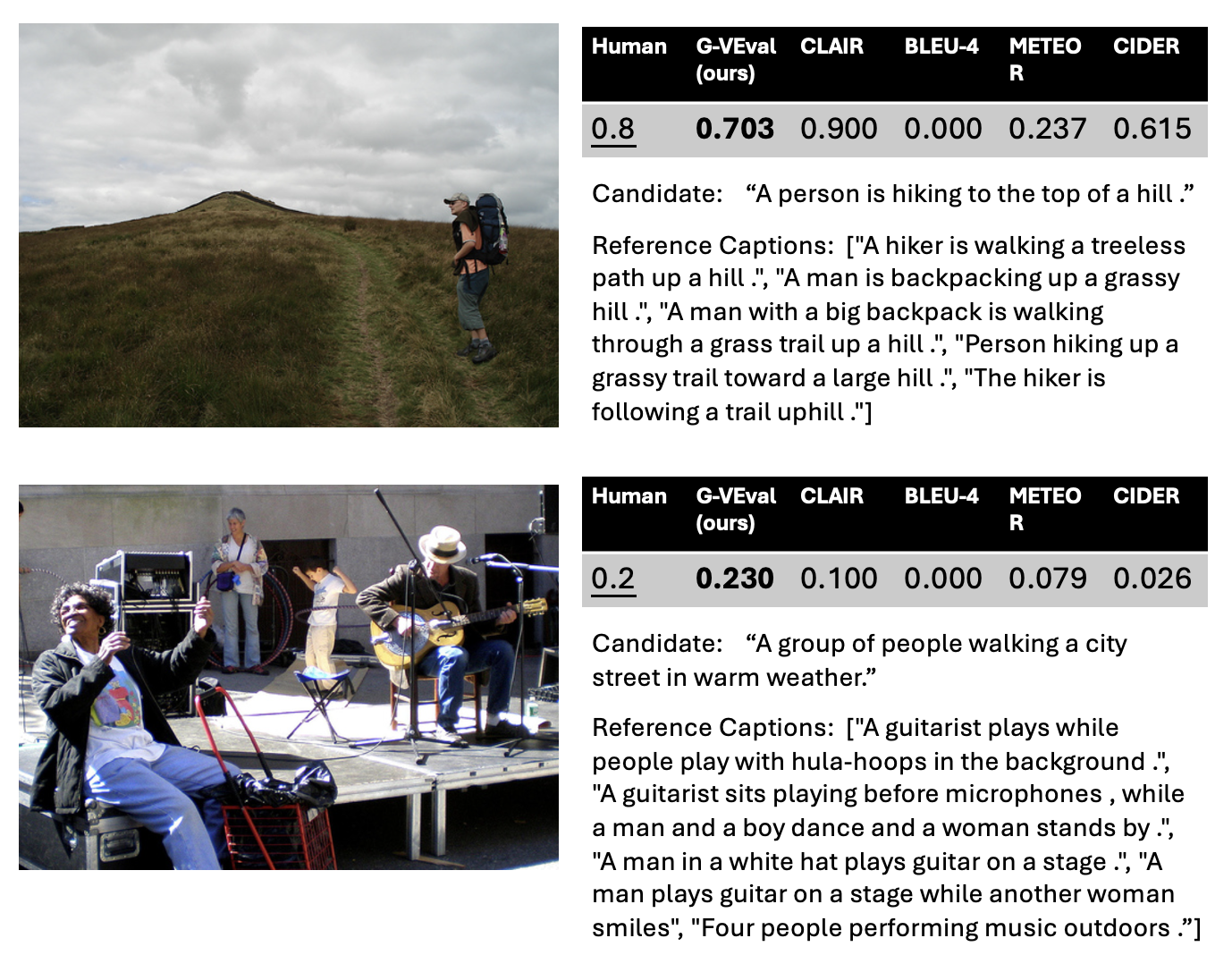}
\caption{Sample qualitative output from the Flickr-8k dataset}
\label{flickr-sample}
\end{figure*}

See Figure \ref{flickr-sample} for G-VEval's evaluation of image captions.

\subsection{VATEX-EVAL}
\begin{figure*}[ht]
\centering
\includegraphics[width=0.95\textwidth]{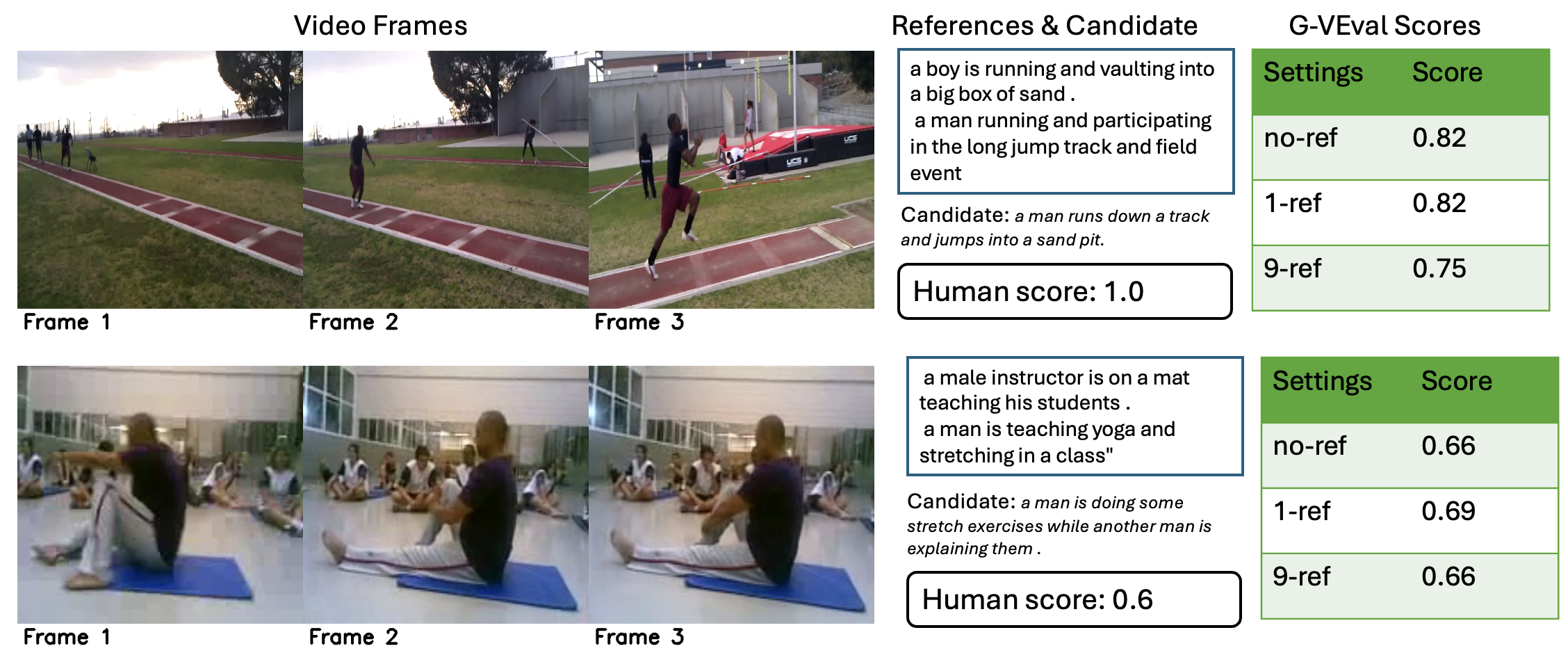}
\caption{Sample qualitative output from the VATEX-EVAL dataset}
\label{vatex-figure}
\end{figure*}

See Figure \ref{vatex-figure} for G-VEval's evaluation of video captions.

\subsection{MSVD-Eval}
\begin{figure*}[ht]
\centering
\includegraphics[width=\textwidth]{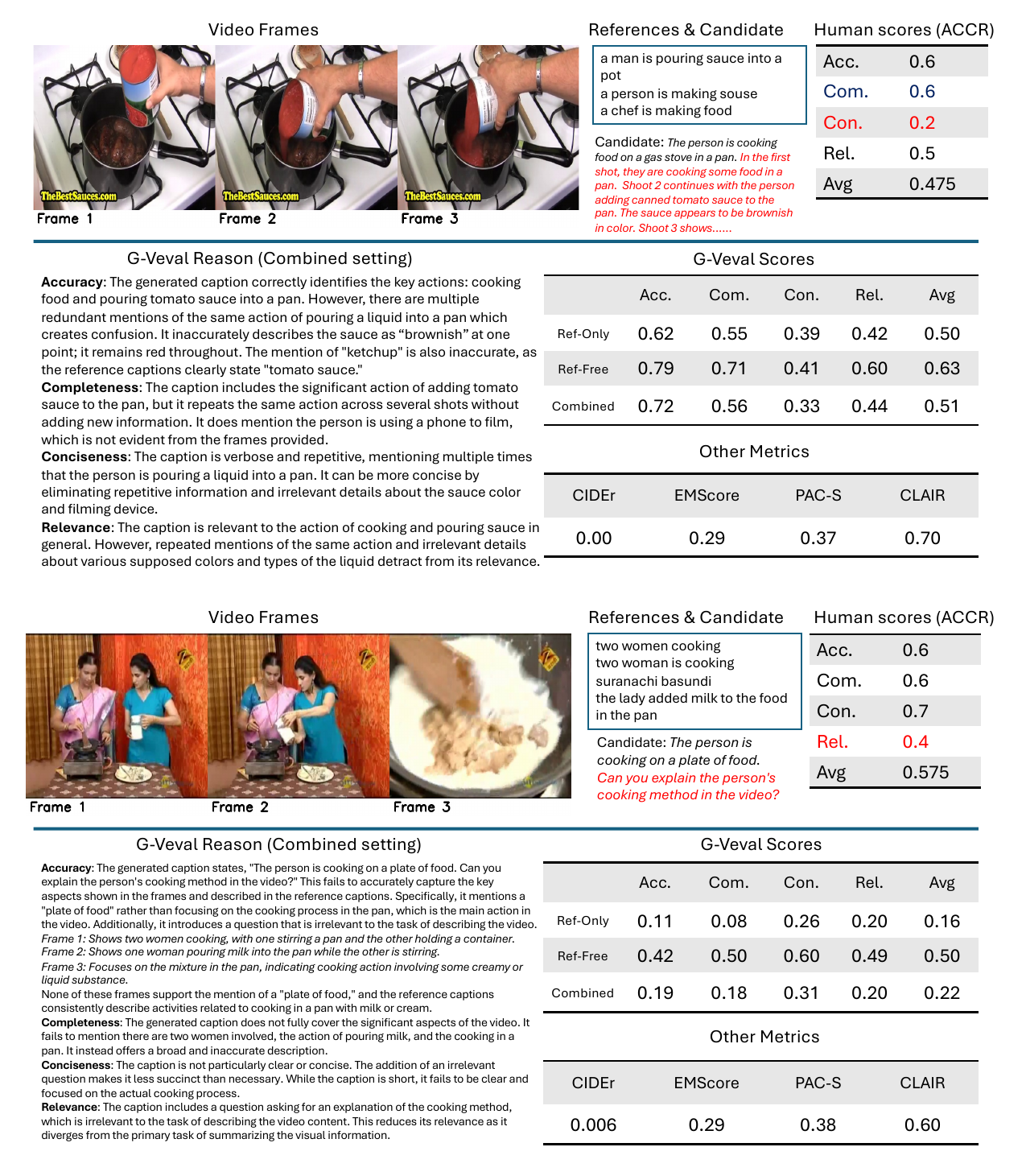}
\caption{Sample qualitative output from the MSVD-Eval dataset (Part 1)}
\label{msvd-sample-1a}
\end{figure*}

\begin{figure*}[ht]
\centering
\includegraphics[width=\textwidth]{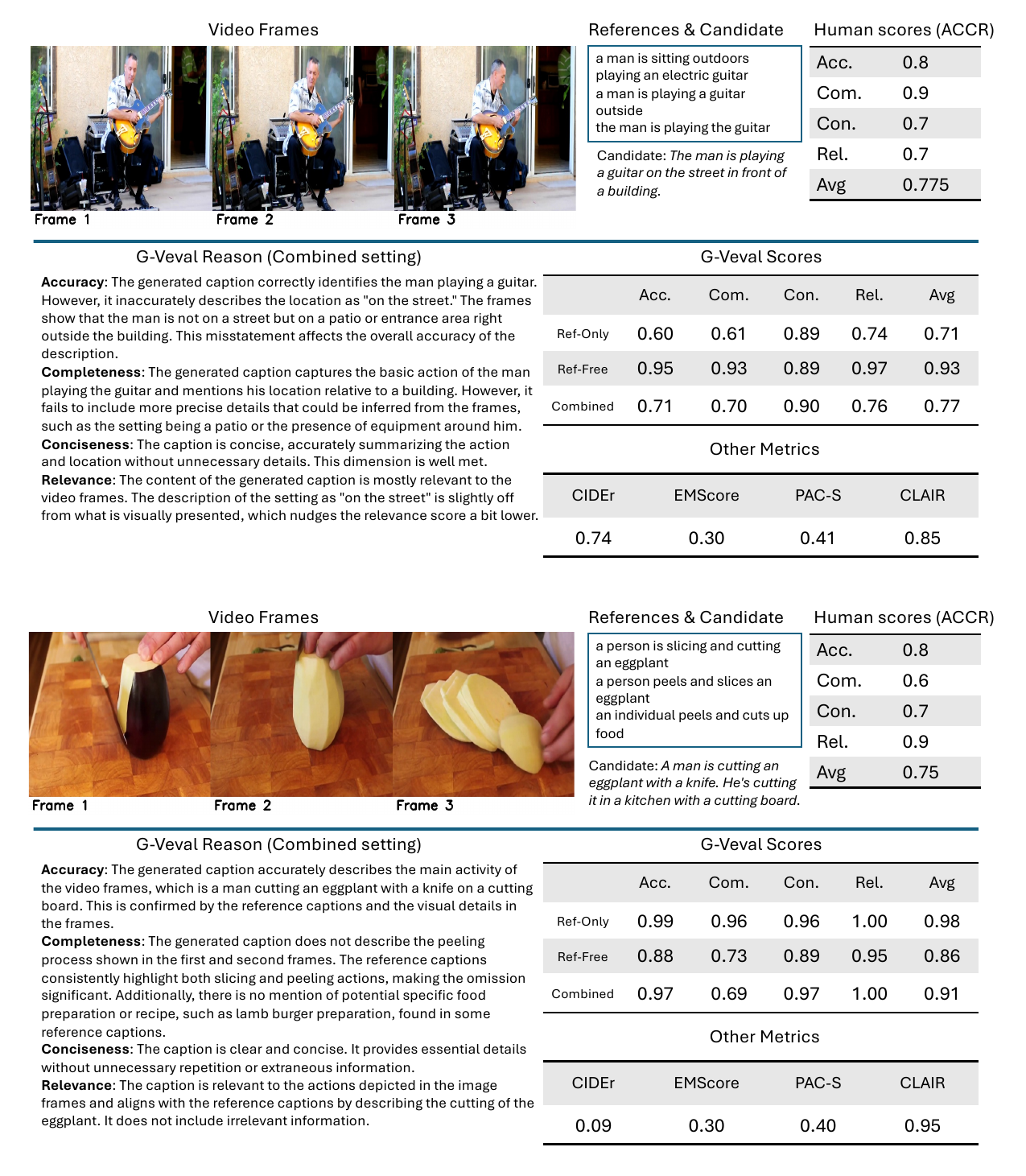}
\caption{Sample qualitative output from the MSVD-Eval dataset (Part 2)}
\label{msvd-sample-1b}
\end{figure*}

See Figure \ref{msvd-sample-1a} and
\ref{msvd-sample-1b} for G-VEval's evaluation results under the ACCR framework.


\section{Appendix E: Supplementary Experiment on FOIL Dataset}

FOIL (Find One mismatch between Image and Language) is a dataset designed to detect hallucinations in captioning models by introducing mismatches between an image and its associated captions. The dataset is used to assess the ability of evaluation metrics to distinguish between accurate and hallucinated captions, making it an important benchmark for evaluating the robustness of captioning models. 

The following table (Table \ref{tab:foil-results}) shows the accuracy of various metrics in the pairwise FOIL hallucination detection setting. All reference-based metrics are given access to either one or four references.

\begin{table}[h!]
\centering
\small
\caption{Accuracy of evaluation metrics in the pairwise FOIL hallucination detection setting.}
\label{tab:foil-results}
\newcommand{\mc}[3]{\multicolumn{#1}{#2}{#3}}
\begin{tabular}{l|cc}
\hline
\textbf{Metric} & \mc{2}{c}{\textbf{FOIL}}  \\
& \textbf{1-ref} & \textbf{4-ref} \\
\hline
BLEU & 66.5 & 82.6 \\
ROUGE & 71.1 & 79.3 \\
METEOR & 78.8 & 82.6 \\
CIDER & 82.5 & 90.6 \\
SPICE & 75.5 & 86.1 \\
\hline
BARTScore & 85.3 & 91.1 \\
MoverScore & 88.4 & 88.4 \\
BERTScore & 88.6 & 92.1 \\
\hline
CLIP-S & 87.2 & 87.2 \\
MID & 90.5 & 90.5 \\
\hline
PAC-S & 89.9 & 89.9 \\
RefCLIP-S & 91.0 & 92.6 \\
RefPAC-S & 93.7 & 94.9 \\
\hline
\textbf{G-VEval} & \textbf{97.8} & \textbf{98.4}  \\
\hline
\end{tabular}
\end{table}

The results on the FOIL benchmark demonstrate that G-VEval achieves state-of-the-art (SOTA) performance in detecting hallucinations in captions. This superior performance indicates G-VEval's robustness and precision in identifying mismatches between visual content and generated captions, further solidifying its effectiveness as a reliable evaluation metric for visual captioning tasks.

\vfill
\textit{(Figures on next page)}


\end{document}